\documentclass[11pt,english]{article}
\usepackage[left=1in,right=1in,top=1.in,bottom=1.in]{geometry}

\usepackage{setspace}
\usepackage{graphicx,graphics}
\usepackage{amssymb}
\usepackage{amsbsy}
\usepackage{stmaryrd} 
\usepackage{amsmath}
\usepackage{multirow}
\usepackage{graphicx}
\usepackage{url}  
\usepackage[toc,page]{appendix}

\usepackage{ragged2e}
\usepackage{dblfloatfix}
\usepackage{booktabs}       

\DeclareMathOperator*{\tr}{tr}

\newtheorem{theorem}{Theorem}

\newtheorem{lemma}{Lemma}
\newtheorem{cor}{Corollary}
\newtheorem{asu}{Assumption}
\newenvironment{Proof}{\paragraph{Proof:}}{\hfill$\square$}
\usepackage[square,numbers]{natbib}

\newcommand{\B}{\boldsymbol}
\usepackage{amsmath,color}

\usepackage[toc,page]{appendix}
\usepackage{times}

\begin{document}
\author{%
    Antoine Dedieu\footnote{Email: \texttt{antoine@vicarious.com}, \texttt{adedieu@mit.edu}. Antoine Dedieu was partially supported by the Office of Naval Research: N000141512342, by MIT and by Vicarious AI. }
    \smallskip
    \\
    \small{\textsc{Vicarious AI, Massachusetts Institute of Technology}}
}
\date{December, 2019}

\title{An error bound for Lasso and Group Lasso in high dimensions}
\maketitle

\begin{abstract}
    We leverage recent advances in high-dimensional statistics to derive new L2 estimation upper bounds for Lasso and Group Lasso in high-dimensions. For Lasso, our bounds scale as $(k^*/n) \log(p/k^*)$---$n\times p$ is the size of the design matrix and $k^*$ the dimension of the ground truth $\B{\beta}^*$---and match the optimal minimax rate. For Group Lasso, our bounds scale as $(s^*/n) \log\left( G / s^* \right) + m^* / n$---$G$ is the total number of groups and $m^*$ the number of coefficients in the $s^*$ groups which contain $\B{\beta}^*$---and improve over existing results. We additionally show that when the signal is strongly group-sparse, Group Lasso is superior to Lasso.
\end{abstract}



\section{Introduction}

We consider the Gaussian linear regression framework, with response $\B{y} \in \mathbb{R}$ and model matrix $\B{X} \in \mathbb{R}^{n \times p}$:
\begin{equation}\label{linreg}
\B{y} = \B{X} \B{\beta}^* + \B{\epsilon}
\end{equation}
where the entries of $\B{\epsilon}=(\epsilon_1, \ldots, \epsilon_p)$ are independent realizations of a sub-Gaussian random variable (as defined in \cite{lecture-notes}) with variance $\sigma^2$. We consider settings where $\B{\beta}^*$ is sparse, i.e. has a small number of non-zeros. The L1-regularized least squares estimator (also known as Lasso estimator \cite{tibshirani1996regression}) is well-known to encourage sparsity in the coefficients. It is defined as a solution of the convex optimization problem:
\begin{equation} \label{l1-problem}
\min \limits_{ \B{\beta} \in \mathbb{R}^{p} } \;\; \frac{1}{n} \| \B{y} - \B{X} \B{\beta}  \|_2^2+ \lambda \| \B{\beta} \|_1.
\end{equation}
In several applications, sparsity is structured---the coefficient indices of $\B{\beta}^*$ occur in groups a-priori  known and it is desirable to select a whole group. In this context, group variants of the L1 regularization are often used to improve the performance and interpretability~\cite{yuan2006model, huang2010benefit}. We consider the Group L1-L2 regularization \cite{bach2011convex} and define the Group Lasso estimator as a solution of the convex problem:
\begin{equation} \label{group-problem}
\min \limits_{ \B{\beta} \in \mathbb{R}^{p} } \;\; \frac{1}{n} \| \B{y} - \B{X} \B{\beta} \|_2^2 + \lambda \sum_{g=1}^G \| \B{\beta}_g \|_{2}
\end{equation}
where, $g=1, \ldots, G$ denotes a group index (the groups are disjoint), $\B\beta_{g}$ denotes the vector of coefficients belonging to group $g$, $\mathcal{I}_g$ the corresponding set of indexes, $n_g = | \mathcal{I}_g|$ and $\B{\beta} = \left(\B{\beta}_1, \ldots, \B{\beta}_G \right)$. In addition, we denote $g_* := \max_{g=1,\ldots,G} n_g$, $\mathcal{J}^* \subset \{1, \ldots, G \}$ the smallest subset of group indexes such that the support of $\B{\beta}^*$ is included in the union of these groups, $s^*:= | \mathcal{J}^* |$ the cardinality of $\mathcal{J}^*$,  and $m^*$ the sum of the sizes of these $s^*$ groups. 



\paragraph{Existing work on statistical performance}
Statistical performance and L2 consistency for high-dimensional linear regression have been widely studied \cite{candes2007dantzig, lasso-dantzig, candes-sparse-estimation, bellec2018slope, lounici2011oracle}. One important statistical performance measure is the L2 estimation error defined as $\| \hat{\B{\beta}} - \B{\beta}^* \|_2$ where $\B{\beta}^*$ is the $k^*$-sparse ground truth used in Equation \eqref{linreg} and $\hat{\B{\beta}}$ is an estimator. For regression problems with least-squares loss, \cite{candes-sparse-estimation} and  \cite{raskutti_wainwright} established a  $(k^*/n)\log(p/k^*)$ lower bound for estimating the L2 norm of a sparse vector, regardless of the input matrix and estimation procedure. This optimal minimax rate is known to be achieved by a global minimizer of a L0 regularized estimator  \cite{bic-tsybakov} which is, however, intractable in practice. Recently, \cite{bellec2018slope} reached this optimal minimax bound for a Lasso estimator---improving over existing results \cite{lasso-dantzig}---and for a recently introduced and tractable Slope estimator. In addition, when sparsity is structured, \cite{huang2010benefit} proved a $(s^*/n) \log(G)+ m^*/n$ L2 estimation upper bound for a Group Lasso estimator---where, similarly to our notations, $G$ is the number of groups, $s^*$ the number of relevant groups and $m^*$ their aggregated size---and showed that this estimator is superior to standard Lasso when the signal is strongly group-sparse, i.e. $m^* / k^*$ is low and the signal is efficiently covered by the groups. \cite{lounici2011oracle} similarly showed that, in the multitask setting, a Group Lasso estimator is superior to Lasso.

\bigskip
\noindent{\bf{What this paper is about:} } In this short paper, we propose a statistical framework to study the L2 estimation performance of Lasso and Group Lasso in high dimensions and we derive new error bounds for these estimators. To this end, we adapt proof techniques recently developed for high-dimensional classification studies \cite{dedieu2019sparse, dedieu2018} to the least squares case. Our bounds are reached under standard assumptions and hold with high probability and in expectation. For Lasso, our bounds scale as $(k^*/n)\log(p/k^*)$: they reach the optimal minimax rate \cite{raskutti_wainwright} while matching the best results \cite{bellec2018slope}. For Group Lasso, our bounds scale as $(s^*/n)\log(G/s^*)+ m^*/ n$ and improve over existing results \cite{huang2010benefit}, due to a stronger cone condition (cf. Theorem \ref{cone-condition}). We additionally recover the result that when the signal is strongly group-sparse, Group Lasso is superior to Lasso.

\section{Statistical analysis}\label{sec: error-bound}

Similarly to the regression literature \cite{lasso-dantzig, bellec2018slope, lounici2011oracle}, we reach our bounds for the Lasso and Group Lasso estimators by assuming a bound on the L2 norm of the columns of $\B{X}$ and restricted eigenvalue conditions.

\subsection{L2 bound on the columns or groups} 
For Lasso, Assumption \ref{asu-col} assumes a standard bound on the L2 norm of the columns of the model matrix. For Group Lasso, Assumption  \ref{asu4}$.3$ assumes an upper bound for the quadratic form associated with $\B{X}_g^T \B{X}_g$ on each group.  
\begin{asu} \label{asu-col}
\begin{itemize}
\item Assumption \ref{asu-col}.1 holds if the model matrix satisfies $\| \B{X}_j \|_2 \le \sqrt{n}, ~ \forall j.$

\item  Let $\B{X}_g \in \mathbb{R}^{n \times n_g}$ denote the restriction of the model matrix to the columns $\mathcal{I}_g$ of group $g$, and let $\mu_{\max}(\B{X}_g^T \B{X}_g)$ be the highest eigenvalue of the positive semi-definite symmetric matrix $\B{X}_g^T \B{X}_g$. Assumption \ref{asu-col}.2 is satisfied if it almost surely holds:
$$ \sup \limits_{g=1, \ldots G} \mu_{\max} (\B{X}_g^T \B{X}_g)  \le n.$$
\end{itemize}
\end{asu}

\subsection{Restricted eigenvalue conditions} Assumptions \ref{asu4}$.1$ and \ref{asu4}$.2$  ensure that the quadratic form associated with the Hessian matrix $n^{-1}\B{X}^T\B{X}$ is lower-bounded on a family of cones of $\mathbb{R}^{p}$---specific to the regularization used. 

\begin{asu} \label{asu4}
    \begin{itemize}
    
    \item  Let $\gamma_1, \gamma_2>0$. Assumption \ref{asu4}.1$(k, \gamma)$ holds if there exists $\kappa(k, \gamma_1, \gamma_2)$ which almost surely satisfies:
    $$0 < \kappa(k, \gamma_1, \gamma_2) \le \inf \limits_{| S | \le k } \ \inf \limits_{ \substack{\B{z} \in \Lambda(S, \gamma_1, \gamma_2) } } \frac{ \B{z}^T  \B{X}^T \B{X} \B{z}    }{ n \|\B{z}\|_2^2  },$$
    where $\gamma=(\gamma_1, \gamma_2)$ and for every subset $S \subset \left\{1, \ldots ,p\right\}$, the cone $\Lambda(S, \gamma_1, \gamma_2) \subset \mathbb{R}^{p}$ is defined as:
    $$\Lambda(S, \gamma_1, \gamma_2) = \left\{ \B{z} \in \mathbb{R}^{p}: \ \| \B{z}_{S^c}  \|_1 \le \gamma_1 \| \B{z}_{S}  \|_1 + \gamma_2 \| \B{z}_{S}  \|_2 \right\}.$$
    
    \item  Let $\epsilon_1, \epsilon_2 > 0$. Assumption \ref{asu4}.2$(s, \epsilon)$ holds if there exists a constant $\kappa(s, \epsilon_1, \epsilon_2)>0$ such that a.s.:
    $$0 < \kappa(s, \epsilon_1, \epsilon_2) \le \inf \limits_{| \mathcal{J} | \le s } \ \inf \limits_{ \substack{\B{z} \in \Omega(\mathcal{J}, \epsilon_1, \epsilon_2) } } \frac{ \B{z}^T \B{X}^T \B{X} \B{z}   }{ n \|\B{z}\|_2^2  },$$
     where $\epsilon=(\epsilon_1, \epsilon_2)$ and for every subset $\mathcal{J} \subset \left\{1, \ldots ,G \right\}$, we define $\mathcal{T}(\mathcal{J}) = \cup_{g \in \mathcal{J}} \mathcal{I}_g$ as the subset of all indexes across all the groups in $\mathcal{J}$. $\Omega(\mathcal{J}, \epsilon_1, \epsilon_2) $ is defined as:
    $$\Omega(\mathcal{J}, \epsilon_1, \epsilon_2) = \left\{ \B{z} \in \mathbb{R}^{p}: \ \sum_{g \notin \mathcal{J}} \| \B{z}_g \|_{2} \le  \epsilon_1 \sum_{g \in \mathcal{J}} \| \B{z}_g \|_{2} + \epsilon_2 \| \B{z}_{\mathcal{T}(\mathcal{J})  } \|_{2} \right\}.$$
    \end{itemize}
\end{asu}

\subsection{Cone conditions}\label{sec: cone condition}

Similarly to existing work~\cite{lasso-dantzig, bellec2018slope, lounici2011oracle}, Theorem \ref{cone-condition} derives a cone condition satisfied by a Lasso or Group Lasso estimator. In particular, Theorem \ref{cone-condition} says that, the difference between the estimator and the ground truth $\B{\beta}^*$ belongs to one of the families of cones defined in Assumption \ref{asu4}. 
The cone conditions are derived by selecting a regularization parameter large enough so that it dominates the gradient of the least squares loss evaluated at the theoretical minimizer $\B{\beta}^*$.

\begin{theorem} \label{cone-condition}
    Let $\delta \in \left(0, \frac{1}{n} \right)$, $\alpha\ge 2$. The following results holds with probability at least $1 - \delta$:
    
    \begin{itemize}
    \item  Let us assume that Assumption \ref{asu-col}.1 holds. Let $\hat{\B{\beta}}_1$ be a solution of the Lasso Problem  \eqref{l1-problem} with parameter $\lambda=24 \alpha  \sigma \sqrt{ n^{-1} \log(2pe/k^*) \log(1 / \delta)}$, and let $S_0\subset \{1,\ldots,p\}$ be the subset of indexes of the $k^*$ highest coefficients of $\B{h}_1:= \hat{\B{\beta}}_1 - \B{\beta}^*$. It holds:
    $$\B{h}_1 \in \Lambda \left( S_0, \ \gamma_1^*:= \frac{\alpha}{\alpha -1},  \ \gamma_2^*:= \frac{\sqrt{k^*}}{\alpha -1} \right).$$
    
    \item Let us assume that Assumption \ref{asu-col}.2 holds. Let $\hat{\B{\beta}}_{L1-L2}$ be a solution of the Group Lasso Problem \eqref{group-problem} with parameter $\lambda_G=24 \alpha\sigma \sqrt{ n^{-1}  \log(2Ge/s^*)\log(2/ \delta)  } + 4\alpha \sigma \sqrt{ n^{-1}  \gamma (s^*)^{-1} m^* } $. Let $\mathcal{J}_0 \subset \{1,\ldots,G\}$ be the subset of indexes of the $s^*$  highest groups of $\B{h}_{L1-L2} :=\hat{\B{\beta}}_{L1-L2}  - \B{\beta}^*$ for the L2 norm. We additionally denote $m_0$ be the total size of the $s^*$ largest groups and assume  $m_0 \le \gamma m^*$ for some $\gamma \ge 1$. It then holds:
    $$\B{h}_{L1-L2} \in \Omega \left( \mathcal{J}_0, \; \epsilon^*_1:=\frac{\alpha}{\alpha - 1}, \; \epsilon^*_2:=\frac{\sqrt{s^*}}{\alpha - 1} \right).$$
    \end{itemize}
\end{theorem}
The proof is presented in Appendix \ref{sec: appendix_cone-condition}: it uses a new result from \cite{dedieu2019sparse} to control the maximum of sub-Gaussian random variables. As a consequence, the regularization parameter $\lambda^2$ for Lasso is of the order of $\log(p/k^*) /n $ and is stronger than existing results \cite{lasso-dantzig}. For Group Lasso, our parameter $\lambda_G^2$ is of the order of $\log(G/s^*) / n + m^* /  (s^* n)$ and improve over \cite{huang2010benefit}---which considers a scaling of $\log(G) / n + m^*/n$.

\subsection{Upper bounds for L2 coefficients estimation}
We now state our main bounds in Theorem \ref{main-results} and Corollary \ref{main-corollary}. 
\begin{theorem} \label{main-results}
    Let $\delta \in \left(0, \frac{1}{2} \right), ~ \alpha \ge 2$. We consider the same assumptions and notations than Theorem \ref{cone-condition}. 
    
    \begin{itemize}
    \item If Assumptions \ref{asu-col}.1 and \ref{asu4}.1$(k^*, \gamma^*)$ hold, the Lasso estimator satisfies with probability at least $1-\delta$:
    \begin{equation*}
    \begin{split}   
    \| \hat{\B{\beta}}_{1} - \B{\beta}^*\|_2 \lesssim &
     \frac{ \alpha \sigma }{\kappa^*} \sqrt{ \frac{ k^* \log\left( p/k^* \right) \log\left( 1/\delta \right) }{n}}.
    \end{split}
    \end{equation*}
    \item If Assumptions \ref{asu-col}.2 and \ref{asu4}.2$(s^*, \epsilon^*)$ holds, the Group Lasso estimator satisfies with same probability:
    \begin{equation*}
    \begin{split}
    &\| \hat{\B{\beta}}_{L1-L2} - \B{\beta}^* \|_2 \lesssim 
    \frac{ \alpha \sigma }{\kappa^*} \sqrt{ \frac{ s^* \log\left( G/s^* \right) \log\left( 1/\delta \right) + \gamma m^* }{n}}.
    \end{split}
    \end{equation*}
        \end{itemize}
    where $\kappa^* = \kappa \left(S_0, \gamma_1^*, \gamma_2^* \right)$ for Lasso and $\kappa^* = \kappa \left(\mathcal{J}_0, \epsilon_1^*, \epsilon_2^*\right)$ for Group Lasso.
\end{theorem}
The proof is presented in Appendix \ref{sec: appendix_main-results}. The bounds directly follow from the the cone conditions proofs and the use of the restricted eigenvalue assumptions. Theorem \ref{main-results} holds for any $\delta \le \frac{1}{2}$. Thus, we obtain by integration the following bounds in expectation. The proof is presented in Appendix \ref{sec: appendix_main-corollary}.
\begin{cor} \label{main-corollary}
    The bounds presented in Theorem \ref{main-results} additionally holds in expectation, that is:
    \begin{equation*}
    \mathbb{E} \| \hat{\B{\beta}}_{1}  - \B{\beta}^*\|_2 \lesssim \frac{\alpha \sigma}{\kappa^*}  \sqrt{ \frac{k^* \log\left( p /k^* \right)}{n} },
    \end{equation*}
    \begin{equation*}
    \mathbb{E} \| \hat{\B{\beta}}_{L1-L2}  - \B{\beta}^*\|_2 \lesssim \frac{\alpha \sigma}{\kappa^*} \sqrt{ \frac{s^* \log\left( G / s^* \right) + \gamma m^*}{n} }.
    \end{equation*}
\end{cor}
\paragraph{Discussion: } For Lasso, our bounds scale as $(k^ / n) \log(p/k^*)$. They improve over \cite{lasso-dantzig} and match the best existing result \cite{bellec2018slope} while reaching the optimal minimax rate. For Group Lasso, our bounds scale as $(s^* / n) \log\left( G/s^* \right) + m^* / n$ and improve over \cite{huang2010benefit}. This is due to the stronger cone condition derived in Theorem \ref{cone-condition}. For both cases, our bounds reach the same scaling than the respective L1 and Group L1-L2 regularizations discussed in \cite{dedieu2019sparse}, which considers a general learning problem with Lipschitz loss functions (including hinge, logistic and quantile regression losses).

\paragraph{Comparison for group-sparse signals: }
We compare the statistical performance and upper bounds for Lasso and Group Lasso when sparsity is structured.
Let us first consider two edge case. \textbf{(i)} If all the groups are all of size $k^*$ and the optimal solution is contained in only one group---that is, $g_*=k^*$, $G = \lceil p / k^* \rceil$, $s^*=1$, $m=k^*$ $\gamma=1$---the  rate for Group Lasso is lower than the one for Lasso. Group Lasso is superior as it strongly exploits the problem structure. \textbf{(ii)} If now all the groups are of size one---that is, $g_*= 1$, $G = p$, $s^*=k^*$, $m^*=k^*$ $\gamma=1$---then Group Lasso has not advantage over Lasso.

\medskip
\noindent
Let us now consider the general case. If $m_* / k_* \ll \log(p/k^*) / \gamma$, then the signal is efficiently covered by the groups---the group structure is useful---and the Group Lasso rate is lower than the Lasso one. That is, similarly to the regression case \cite{huang2010benefit}, Group Lasso is superior to Lasso for strongly group-sparse signals. However, when $m_* / k_*$ is larger, then sparsity is not as useful and Group Lasso is outperformed by Lasso.

\newpage
\bibliographystyle{plainnat}
\bibliography{regression}

\begin{thebibliography}{15}
\providecommand{\natexlab}[1]{#1}
\providecommand{\url}[1]{\texttt{#1}}
\expandafter\ifx\csname urlstyle\endcsname\relax
  \providecommand{\doi}[1]{doi: #1}\else
  \providecommand{\doi}{doi: \begingroup \urlstyle{rm}\Url}\fi

\bibitem[Bach et~al.(2011)Bach, Jenatton, Mairal, and
  Obozinski]{bach2011convex}
Francis Bach, Rodolphe Jenatton, Julien Mairal, and Guillaume Obozinski.
\newblock Convex optimization with sparsity-inducing norms.
\newblock \emph{Optimization for Machine Learning}, 5:\penalty0 19--53, 2011.

\bibitem[Bellec et~al.(2018)Bellec, Lecu{\'e}, Tsybakov,
  et~al.]{bellec2018slope}
Pierre~C Bellec, Guillaume Lecu{\'e}, Alexandre~B Tsybakov, et~al.
\newblock Slope meets {Lasso}: improved oracle bounds and optimality.
\newblock \emph{The Annals of Statistics}, 46\penalty0 (6B):\penalty0
  3603--3642, 2018.

\bibitem[Bickel et~al.(2009)Bickel, Ritov, and Tsybakov]{lasso-dantzig}
Peter~J Bickel, Ya'acov Ritov, and Alexandre~B Tsybakov.
\newblock Simultaneous analysis of {Lasso} and {Dantzig} selector.
\newblock \emph{The Annals of Statistics}, pages 1705--1732, 2009.

\bibitem[Bunea et~al.(2007)Bunea, Tsybakov, Wegkamp, et~al.]{bic-tsybakov}
Florentina Bunea, Alexandre~B Tsybakov, Marten~H Wegkamp, et~al.
\newblock Aggregation for {Gaussian} regression.
\newblock \emph{The Annals of Statistics}, 35\penalty0 (4):\penalty0
  1674--1697, 2007.

\bibitem[Candes and Davenport(2013)]{candes-sparse-estimation}
Emmanuel Candes and Mark~A Davenport.
\newblock How well can we estimate a sparse vector?
\newblock \emph{Applied and Computational Harmonic Analysis}, 34\penalty0
  (2):\penalty0 317--323, 2013.

\bibitem[Candes and Tao(2007)]{candes2007dantzig}
Emmanuel~J Candes and Terence Tao.
\newblock The {Dantzig} selector: statistical estimation when p is much larger
  than n.
\newblock \emph{The Annals of Statistics}, pages 2313--2351, 2007.

\bibitem[Dedieu(2019{\natexlab{a}})]{dedieu2018}
Antoine Dedieu.
\newblock Error bounds for sparse classifiers in high-dimensions.
\newblock \emph{Proceedings of Machine Learning Research}, 89:\penalty0 48--56,
  2019{\natexlab{a}}.

\bibitem[Dedieu(2019{\natexlab{b}})]{dedieu2019sparse}
Antoine Dedieu.
\newblock Sparse (group) learning with lipschitz loss functions: a unified
  analysis.
\newblock \emph{arXiv preprint arXiv:1910.08880}, 2019{\natexlab{b}}.

\bibitem[Hsu et~al.(2012)Hsu, Kakade, Zhang, et~al.]{hsu2012tail}
Daniel Hsu, Sham Kakade, Tong Zhang, et~al.
\newblock A tail inequality for quadratic forms of subgaussian random vectors.
\newblock \emph{Electronic Communications in Probability}, 17, 2012.

\bibitem[Huang and Zhang(2010)]{huang2010benefit}
Junzhou Huang and Tong Zhang.
\newblock The benefit of group sparsity.
\newblock \emph{The Annals of Statistics}, 38\penalty0 (4):\penalty0
  1978--2004, 2010.

\bibitem[Lounici et~al.(2011)Lounici, Pontil, Van De~Geer, Tsybakov,
  et~al.]{lounici2011oracle}
Karim Lounici, Massimiliano Pontil, Sara Van De~Geer, Alexandre~B Tsybakov,
  et~al.
\newblock Oracle inequalities and optimal inference under group sparsity.
\newblock \emph{The Annals of Statistics}, 39\penalty0 (4):\penalty0
  2164--2204, 2011.

\bibitem[Raskutti et~al.(2011)Raskutti, Wainwright, and
  Yu]{raskutti_wainwright}
Garvesh Raskutti, Martin~J Wainwright, and Bin Yu.
\newblock Minimax rates of estimation for high-dimensional linear regression
  over $l_q$-balls.
\newblock \emph{IEEE transactions on information theory}, 57\penalty0
  (10):\penalty0 6976--6994, 2011.

\bibitem[Rigollet(2015)]{lecture-notes}
Philippe Rigollet.
\newblock 18.s997: High dimensional statistics.
\newblock \emph{Lecture Notes), Cambridge, MA, USA: MIT OpenCourseWare}, 2015.

\bibitem[Tibshirani(1996)]{tibshirani1996regression}
Robert Tibshirani.
\newblock Regression shrinkage and selection via the {Lasso}.
\newblock \emph{Journal of the Royal Statistical Society. Series B
  (Methodological)}, pages 267--288, 1996.

\bibitem[Yuan and Lin(2006)]{yuan2006model}
Ming Yuan and Yi~Lin.
\newblock Model selection and estimation in regression with grouped variables.
\newblock \emph{Journal of the Royal Statistical Society: Series B (Statistical
  Methodology)}, 68\penalty0 (1):\penalty0 49--67, 2006.

\end{thebibliography}

\newpage
\begin{appendices}

\section {Proof of Theorem \ref{cone-condition}}  \label{sec: appendix_cone-condition}
We use the minimality of  $\hat{\B{\beta}}$ and Lemma 4 from \cite{dedieu2019sparse} to derive the cone conditions:
\begin{lemma}\label{upper-bound-sup} 
    \textbf{(Lemma 4, \cite{dedieu2019sparse})}
    Let $g_1,\ldots g_r$ be sub-Gaussian random variables with variance $\sigma^2$. We denote by $(g_{(1)}, \ldots, g_{(r)})$ a non-increasing rearrangement of $(|g_1|, \ldots, |g_r|)$ and define the coefficients $\lambda_j^{(r)} = \sqrt{ \log(2r/j) }, \ j=1,\ldots\,r$. For $\delta \in \left(0, \frac{1}{n} \right)$, it holds with probability at least $1-\delta$:
    $$ \sup \limits_{j=1,\ldots,r} \left\{ \frac{ g_{(j)} }{\sigma \lambda_j^{(r)}} \right\} \le 12 \sqrt{ \log(1 / \delta)}.  $$ 
\end{lemma}

\begin{Proof}  We first present the proof for the Lasso estimator before adapting it to Group Lasso.
\medskip
\noindent
 \textbf{Proof for Lasso: } $\hat{\B{\beta}}$ denotes herein the Lasso estimator. $\hat{\B{\beta}}$ is a solution of the Lasso Problem \eqref{group-problem} hence:
    \begin{align*}
    \begin{split}
    \frac{1}{n}  \|  \B{y} - \B{X} \hat{\B{\beta}} \|_2^2 + \lambda \| \hat{\B{\beta}} \|_1
    &\le \frac{1}{n}  \| \B{y} - \B{X} \B{\beta}^*  \|_2^2 + \lambda \| \B{\beta}^* \|_1 = \frac{1}{n}  \| \B{\epsilon}   \|_2^2 + \lambda \| \B{\beta}^* \|_1.
    \end{split}
    \end{align*}
    Since we have defined $\B{h} =  \hat{\B{\beta}} - \B{\beta}^*$, it holds:
    $$ \frac{1}{n} \|  \B{y} - \B{X} \hat{\B{\beta}} \|_2^2 
    =  \frac{1}{n} \|  \B{X} \B{\beta}^*  - \B{X} \hat{\B{\beta}} \|_2^2 + \frac{2}{n} \B{\epsilon}^T (\B{X} \B{\beta}^*  - \B{X} \hat{\B{\beta}}) +   \frac{1}{n} \| \B{\epsilon}   \|_2^2 =  \frac{1}{n} \|  \B{X} \B{h} \|_2^2 - \frac{2}{n} ( \B{X}^T \B{\epsilon})^T \B{h} +   \frac{1}{n} \| \B{\epsilon}   \|_2^2.$$
    Since $S^*$ is the support of $\B{\beta}^*$ and $S_0 = \left\{1,\ldots,k^*\right\}$ is the set of the $k^*$ largest coefficients of $\B{h}$, it holds:
    \begin{align}\label{rhs-lasso}
    \begin{split}
    \frac{1}{n} \| \B{X} \B{h} \|_2^2  
    &\le \frac{2}{n}  ( \B{X}^T \B{\epsilon})^T \B{h} + \lambda   \| \B{\beta}^*_{S^* } \|_1 - \lambda \| \hat{\B{\beta}}_{S^*} \|_1 - \lambda \| \hat{\B{\beta}}_{(S^*)^c } \|_1 \\
    &\le \frac{2}{n} ( \B{X}^T \B{\epsilon})^T \B{h} + \lambda \| \B{h}_{S^*} \|_{1} - \lambda \| \B{h}_{ (S^*)^c} \|_1\\
    &\le \frac{2}{n} ( \B{X}^T \B{\epsilon})^T \B{h} + \lambda \| \B{h}_{S_0 } \|_{1} - \lambda \| \B{h}_{(S_0)^c} \|_1.
    \end{split}
    \end{align}
    We now upper-bound the quantity $( \B{X}^T \B{\epsilon})^T \B{h} $. To this end, we denote $\B{g} = \B{X}^T \B{\epsilon}$.  The entries of $\B{\epsilon}$ are independent, hence Assumption \ref{asu-col}.1 guarantees that $\forall j, g_j$ is sub-Gaussian with variance $n \sigma^2$. In addition, we introduce a non-increasing rearrangement $(g_{(1)}, \ldots, g_{(p)})$  of $(|g_1|, \ldots, |g_p|)$. We assume without loss of generality that $|h_1| \ge \ldots \ge |h_p|$. Lemma \ref{upper-bound-sup} gives, with probability at least $1-\delta$:
    \begin{align}\label{upper-bound-SG-lasso}
    \begin{split}
    ( \B{X}^T \B{\epsilon})^T \B{h}
    &= \sum_{j=1}^p g_j h_j \le \sum_{j=1}^p | g_j | |  h_j | 
    = \sum_{j=1}^p \frac{g_{(j)}}{\sqrt{n}\sigma \lambda_j}  \sqrt{n}\sigma \lambda_j | h_{(j)} | \\
    &\le \sqrt{n}\sigma \sup_{j=1,\ldots,p} \left\{\frac{g_{(j)}}{ \sqrt{n} \sigma \lambda_j }  \right\}   \sum_{j=1}^p \lambda_j  |h_{(j)} |
    \le 12 \sqrt{n} \sigma \sqrt{\log(1/ \delta)} \sum_{j=1}^p \lambda_j  |h_{(j)} |\text{ with Lemma \ref{upper-bound-sup}}\\
    &\le 12 \sqrt{n} \sigma \sqrt{\log(1/ \delta)} \sum_{j=1}^p \lambda_j  |h_{j} | \text{ since } \lambda_1 \ge \ldots \ge \lambda_p \text { and }  | h_1 | \ge \ldots \ge |h_p| \\
    &\le 12 \sqrt{n} \sigma \sqrt{\log(1/ \delta)} \left( \sum_{j=1}^{k^*} \lambda_j  |h_{j} | + \lambda_{k^*} \| \B{h}_{(S_0)^c} \|_1  \right).
    \end{split}
    \end{align}
    Cauchy-Schwartz inequality leads to:
    \begin{align*}
    \sum_{j=1}^{k^*} \lambda_j | h_j |  &\le \sqrt{\sum_{j=1}^{k^*} \lambda_j ^2 } \| \B{h}_{S_0}  \|_2 \le \sqrt{k^*\log(2pe /k^*)}  \| \B{h}_{S_0}  \|_2,
    \end{align*}
    where we have used Stirling formula to obtain
    \begin{align*}
    \sum_{j=1}^{k^*} \lambda_j ^2 = \sum_{j=1}^{k^*}  \log(2p/j) &= k^*  \log(2p) - \log(k^* !) 
    \le k^* \log(2p) - k^*\log(k^*/ e) = k^* \log(2pe /k^*).
    \end{align*}
    Theorem \ref{cone-condition} defines $\lambda = 24 \alpha  \sigma \sqrt{ n^{-1} \log(2pe/k^*) \log(1 / \delta)}$. Because $\lambda_{k^*} \le  \sqrt{\log(2pe/k^*)}$, we can pair Equations \eqref{rhs-lasso} and \eqref{upper-bound-SG-lasso}  to obtain with probability at least $1 - \delta$:
    \begin{align}\label{fundamental-lasso}
    \begin{split}
    \frac{1}{n} \| \B{X} \B{h} \|_2^2 
    &\le \frac{2}{n} ( \B{X}^T \B{\epsilon})^T \B{h} + \lambda \| \B{h}_{S_0} \|_1 -\lambda \| \B{h}_{(S_0)^c}  \|_1 \\
    &\le 24 \frac{\sigma}{ \sqrt{n} }  \sqrt{\log(2pe/k^*) \log(1/ \delta)} \left( \sqrt{k^*}  \| \B{h}_{S_0}  \|_2  +  \| \B{h}_{(S_0)^c}  \|_1 \right)+ \lambda \| \B{h}_{S_0} \|_1 -\lambda \| \B{h}_{(S_0)^c}  \|_1\\
    &= \frac{\lambda}{\alpha} \left( \sqrt{k^*}  \| \B{h}_{S_0}  \|_2  +  \| \B{h}_{(S_0)^c}  \|_1 \right) + \lambda \| \B{h}_{S_0} \|_1 -\lambda \| \B{h}_{(S_0)^c}  \|_1. 
    \end{split}
    \end{align}
    As a first consequence, Equation \eqref{fundamental-lasso} implies that with probability at least $1 - \delta$:
    $$ \lambda \| \B{h}_{(S_0)^c}  \|_1 -  \frac{\lambda}{\alpha} \| \B{h}_{(S_0)^c}  \|_1  \le \lambda \| \B{h}_{S_0} \|_1 + \frac{\lambda}{\alpha} \sqrt{k^*}\| \B{h}_{S_0}  \|_2,$$
    which is equivalent from saying that with probability at least $1 - \delta$:
    $$ \| \B{h}_{(S_0)^c}  \|_1 \le \frac{\alpha}{\alpha -1}  \| \B{h}_{S_0}  \|_1 + \frac{ \sqrt{k^*}}{\alpha -1}  \| \B{h}_{S_0}  \|_2.$$
    We conclude that $\B{h} \in \Lambda \left(S_0, \ \frac{\alpha}{\alpha -1}, \  \frac{\sqrt{k^*}}{\alpha -1} \right)$ with probability at least $1-\delta$.
    
    \medskip
    
    \paragraph{Proof for Group Lasso: } $\hat{\B{\beta}}$ designs herein the Group Lasso estimator. $\hat{\B{\beta}}$ is a solution of the Group Lasso Problem \eqref{group-problem} hence:
    \begin{align*}
    \begin{split}
    \frac{1}{n}  \|  \B{y} - \B{X} \hat{\B{\beta}} \|_2^2 + \lambda_G \sum_{g=1}^G \| \hat{\B{\beta}}_g \|_2 
     &\le \frac{1}{n}  \| \B{y} - \B{X} \B{\beta}^*  \|_2^2 + \lambda_G \sum_{g=1}^G \| \B{\beta}^*_g \|_2 = \frac{1}{n}  \| \B{\epsilon}   \|_2^2 + \lambda_G  \sum_{g=1}^G \| \B{\beta}^*_g \|_2.
    \end{split}
    \end{align*}
    By definition, the support of $\B{\beta}^* $ is included in $\mathcal{J}^*$ and $\mathcal{J}_0 \subset \{1, \ldots, G\}$ is the subset of indexes of the $s^*$ highest groups of $\B{h}$ for the L2 norm. It then holds:
    \begin{align}\label{rhs-group}
    \begin{split}
    \frac{1}{n} \| \B{X} \B{h} \|_2^2  
    &\le \frac{2}{n} ( \B{X}^T \B{\epsilon})^T \B{h} + \lambda_G  \sum_{g=1}^G \| \B{\beta}^*_g \|_2 - \lambda_G \sum_{g=1}^G \| \hat{\B{\beta}}_g \|_2
    = \frac{2}{n} ( \B{X}^T \B{\epsilon})^T \B{h} + \lambda_G  \sum_{g \in \mathcal{J}^*} \| \B{\beta}^*_g \|_2 - \lambda_G \sum_{g=1}^G \| \hat{\B{\beta}}_g \|_2\\
    &\le \frac{2}{n} ( \B{X}^T \B{\epsilon})^T \B{h} + \lambda_G \sum_{g \in \mathcal{J}_0} \| \B{h}_g \|_{2} - \lambda_G \sum_{g \notin \mathcal{J}_0} \| \B{h}_g \|_{2}.
    \end{split}
    \end{align}
    We now upper-bound the quantity $( \B{X}^T \B{\epsilon})^T \B{h} $---again, we denote $\B{g} = \B{X}^T \B{\epsilon}$. Applying Cauchy-Schwartz inequality on each group gives:
    \begin{align}\label{equations-group}
    ( \B{X}^T \B{\epsilon})^T \B{h}
    \le \left|  \langle  \B{g},   \B{h}  \rangle  \right|
    \le \sum_{g=1}^G \left| \langle  \B{g}_g, \B{h}_g  \rangle  \right|
    \le \sum_{g=1}^G \| \B{g}_g \|_2 \|  \B{h}_g  \|_2,
    \end{align}
     Let us fix $g \le G$. We have denoted $n_g$ the cardinality of the set of indexes $\mathcal{I}_g$ of group $g$. It then holds $\forall \B{u}_g \in \mathbb{R}^{n_g}$:
    \begin{align}\label{required-hsu}
    \begin{split}
    \mathbb{E}\left( \exp \left( \B{g}_g^T \B{u}_g \right) \right) 
    &=\mathbb{E}\left( (\B{X}^T \epsilon)_g^T  \B{u}_g \right)
    =\mathbb{E}\left( (\B{X}_g^T \epsilon)^T  \B{u}_g \right) 
    =\mathbb{E}\left( \epsilon^T  \B{X}_g \B{u}_g \right)\\ 
    &=\prod_{i=1}^{n_g} \mathbb{E}\left( \epsilon_i  (\B{X}_g \B{u}_g)_i \right) \text{ by independence}\\ 
    &\le \prod_{i=1}^{n_g} \exp \left( 4 \sigma^2 (\B{X}_g \B{u}_g)_i^2  \right)\text{with Lemma 1.4 from \cite{lecture-notes}}\\ 
    & ~~~~~~~ \left(\text{ since } \forall i, \epsilon_i  (\B{X}_g \B{u}_g)_i \text{ is sub-Gaussian with variance } \sigma^2 (\B{X}_g \B{u}_g)_i^2\right) \\
    &= \exp \left( 4 \sigma^2 \| \B{X}_g \B{u}_g \|_2^2  \right)
    = \exp \left( 4 \sigma^2  \B{u}_g^T \B{X}_g^T \B{X}_g  \B{u}_g   \right)\\ 
    &\le \exp \left( 4 n \sigma^2  \| \B{u}_g  \|_2^2 \right) \text{ since } \mu_{\max}(\B{X}_g^T \B{X}_g ) \le n \text{ with Assumption } \ref{asu-col}.2.\\  
    \end{split}
    \end{align}
    We can then use Theorem 2.1 from \cite{hsu2012tail}. By denoting $\B{I}_g$ the identity matrix of size $n_g$ it holds:
    $$\mathbb{P}\left( \| \B{I}_g \B{g}_g  \|_2^2 \ge 8 n \sigma^2 \left( \tr(\B{I_g}) + 2 \sqrt{\tr(\B{I_g}^2)  t} + 2 ||| \B{I_g} ||| \right) \right) \le e^{-t}, \forall t > 0$$
    which can be equivalently expressed as
    $$\mathbb{P}\left( \| \B{g}_g \|_2^2 \ge 8 n \sigma^2 \left( \sqrt{n_g} + \sqrt{2 t} \right)^2 \right) \le e^{-t}, \forall t > 0$$
    or, with a different formulation:
    $$\mathbb{P}\left( \frac{1}{ \sqrt{n}} \| \B{g}_g \|_2 -2 \sqrt{2}  \sigma \sqrt{n_g}  \ge 4 \sigma \sqrt{t} \right) \le e^{-t}, \forall t > 0$$
    which is equivalent from saying that:
    \begin{equation}\label{subGauss-group}
    \mathbb{P}\left( \frac{1}{ \sqrt{n}} \| \B{g}_g \|_2^2 - 2 \sqrt{2} \sigma\sqrt{n_g}  \ge t \right)  \le \exp\left( \frac{-t^2}{16 \sigma^2} \right), \forall t > 0.
    \end{equation}
    Let us define the random variables
    $f_g = \max \left(0,  \frac{1}{ \sqrt{n}}  \| \B{g}_g \|_2 - 2 \sqrt{2} \sigma \sqrt{n_g} \right), \ g=1,\ldots,G$. Under Equation \eqref{subGauss-group}, $f_g$ satisfies the same tail condition than a sub-Gaussian random variable with variance $8 \sigma^2$ and we can apply Lemma \ref{upper-bound-sup}. 
    
    \smallskip
    \noindent
    To this end, we introduce a non-increasing rearrangement $(f_{(1)}, \ldots, f_{(G)})$  of $(|f_1|, \ldots, |f_G|)$ and a permutation $\psi$ such that $n_{\psi(1)} \ge \ldots \ge n_{\psi(G)}$---where we have defined the group sizes $n_1, \ldots, n_G$. In addition, we assume without loss of generality that $\| \B{ h }_1 \|_{2} \ge \ldots \| \B{ h }_G \|_{2}$ and we note the coefficients $\lambda_g^{(G)} = \sqrt{\log(2Ge / g)}$.
    
    \smallskip
    \noindent
    Following Equation \eqref{equations-group}, we obtain with probability at least $1-\delta$:
    
    \begin{align}\label{upper-bound-SG-group}
    \begin{split}
    \frac{1}{ \sqrt{n}} ( \B{X}^T \B{\epsilon})^T \B{h}
    &\le \sum_{g=1}^G \frac{1}{ \sqrt{n}} \| \B{g}_g \|_2 \|  \B{h}_g  \|_2=\sum_{g=1}^G \left( \frac{1}{ \sqrt{n}} \| \B{g}_g \|_2 - 2 \sqrt{2} \sigma  \sqrt{n_g}  \right) \| \B{h}_g  \|_2 + 2 \sqrt{2} \sigma \sum_{g=1}^G  \sqrt{n_g} \| \B{h}_g  \|_2\\
    &\le \sum_{g=1}^G |f_{g} |  \|  \B{h}_g ||_2 + 2 \sqrt{2} \sigma  \sum_{g=1}^G \sqrt{n_g} \|  \B{h}_g  \|_2\\
    &= \sum_{g=1}^G \frac{f_{(g)}}{2 \sqrt{2}\sigma \lambda_g^{(G)} }   2 \sqrt{2}\sigma \lambda_g^{(G)} \|  \B{h}_{(g)} ||_2 + 2 \sqrt{2} \sigma\sum_{g=1}^G \sqrt{n_g} \|  \B{h}_g  \|_2\\
    &\le \sup_{g=1,\ldots,G} \left\{\frac{f_{(g)}}{2 \sqrt{2}\sigma \lambda_g^{(G)} }  \right\}    \sum_{g=1}^G 2 \sqrt{2}\sigma \lambda_g^{(G)} \|  \B{h}_{(g)} ||_2 + 2 \sqrt{2} \sigma\sum_{g=1}^G \sqrt{n_g} \|  \B{h}_g  \|_2\\
    &\le 24 \sqrt{2} \sigma \sqrt{\log(1/ \delta)} \sum_{g=1}^G \lambda_g^{(G)} \| \B{h}_{(g)} \|_{2} + 2 \sqrt{2} \sigma  \sum_{g=1}^G \sqrt{n_{g}} \|  \B{h}_{g}  \|_2 \text{ with Lemma \ref{upper-bound-sup}}\\
    &\le 34 \sigma \sqrt{\log(1/ \delta)} \sum_{g=1}^G \lambda_g^{(G)} \| \B{h}_g \|_{2} + 2 \sqrt{2} \sigma  \sum_{g=1}^G \sqrt{n_{\psi(g)}} \|  \B{h}_{g}  \|_2 \\
    &~~~~~~\text{ since } \lambda_1^{(G)} \ge \ldots \ge \lambda_G^{(G)},  \| \B{h}_{1} \|_{2} \ge \ldots \ge \| \B{h}_{G} \|_{2} ~\text{ and }~ n_{\psi(1)} \ge \ldots \ge n_{\psi(G)}\\
    &\le 34 \sigma \sqrt{ \log(1/ \delta) } \left(  \sqrt{s^*\log(2Ge /s^*)}   \left( \sum_{g \in \mathcal{J}_0} \| \B{h}_{g} \|_{2}^2 \right)^{1/2} + \lambda_{s^*}^{(G)} \sum_{g \notin \mathcal{J}_0} \| \B{h}_{g} \|_{2} \right)\\
     &~~~+ 4\sigma \left( \left( \sum_{g=1}^{s^*} n_{\psi(g)} \right)^{1/2} \left( \sum_{g \in \mathcal{J}_0} \| \B{h}_{g} \|_{2}^2 \right)^{1/2}  + \max_{g= s^* + 1, \ldots, G} \sqrt{n_{\psi(g)}} \sum_{g \notin \mathcal{J}_0 } \| \B{h}_{g} \|_{2} \right)\\ 
    &\le 34 \sigma \sqrt{ \log(2Ge /s^*) \log(1/ \delta) } \left(  \sqrt{s^*}   \| \B{h}_{\mathcal{T}_0} \|_{2} + \sum_{g \notin \mathcal{J}_0} \| \B{h}_{g} \|_{2} \right) \\
    &~~~+ 4 \sigma \left( \sqrt{m_0} \| \B{h}_{\mathcal{T}_0} \|_{2}  + \sqrt{\frac{m_0}{s^*}} \sum_{g \notin \mathcal{J}_0 } \| \B{h}_{g} \|_{2} \right)\\ 
    &\le \left(34 \sigma \sqrt{ \log(2Ge /s^*) \log(1/ \delta) } + 4 \sigma \sqrt{ \gamma m^* / s^* } \right) \left(  \sqrt{s^*}   \| \B{h}_{\mathcal{T}_0} \|_{2} + \sum_{g \notin \mathcal{J}_0} \| \B{h}_{g} \|_{2} \right),
    \end{split}
    \end{align}
    where we define $\mathcal{T}_0 = \cup_{g \in \mathcal{J}_0} \mathcal{I}_g$ as the subset  of all indexes across all the $s^*$ groups in $\mathcal{J}_0$, and $m_0$ denotes the total size of the $s^*$ largest groups. Note that we have paired Cauchy-Schwartz inequality with Stirling formula to obtain
    $$\sum_{g=1}^{s^*} \left(\lambda_g^{(G)} \right)^2  \le s^* \log(2Ge /s^*).$$
    Theorem \ref{cone-condition} defines  $\lambda_G = 34 \alpha \sigma  \sqrt{ n^{-1}  \log(2Ge/s^*) \log(1/ \delta)} + 4 \alpha \sigma \sqrt{ n^{-1} \gamma (s^*) ^{-1} m^* }$. By pairing Equations  \eqref{rhs-group} and \eqref{upper-bound-SG-group} it holds with probability at least $1 - \delta$:
    \begin{equation}\label{fundamental-group}
    \begin{split}
    \frac{1}{n} \| \B{X} \B{h} \|_2^2  
    &\le \frac{\lambda_G}{\alpha}\left(  \sqrt{s^*}   \| \B{h}_{\mathcal{T}_0} \|_{2} + \sum_{g \notin \mathcal{J}_0} \| \B{h}_{g} \|_{2} \right) + \lambda_G \sum_{g \in \mathcal{J}_0} \| \B{h}_g \|_{2} - \lambda_G \sum_{g \notin \mathcal{J}_0} \| \B{h}_g \|_{2},
    \end{split}
    \end{equation}
    As a first consequence, Equation \eqref{fundamental-group} implies that with probability at least $1 -  \delta$
    \begin{equation*} 
    \begin{split}
     \lambda_G \sum_{g \notin \mathcal{J}_0} \| \B{h}_g \|_{2} - \frac{\lambda_G}{\alpha} \sum_{g \notin \mathcal{J}_0} \| \B{h}_{g} \|_{2} 
    &\le \lambda_G \sum_{g \in \mathcal{J}_0} \| \B{h}_g \|_{2} + \frac{\lambda_G}{\alpha}   \sqrt{s^*}   \| \B{h}_{\mathcal{T}_0} \|_{2} 
    \end{split}
    \end{equation*}
    which is equivalent to saying that with probability at least $1- \delta$:
    \begin{equation*} 
    \sum_{g \notin \mathcal{J}_0} \| \B{h}_g \|_{2}   \le  \frac{\alpha}{\alpha - 1}  \sum_{g \in \mathcal{J}_0} \| \B{h}_g \|_{2} +  \frac{\sqrt{s^*} }{\alpha - 1}  \| \B{h}_{\mathcal{T}_0} \|_{2},
    \end{equation*}
    that is $\B{h} \in \Omega\left(\mathcal{J}_0, \frac{\alpha}{\alpha-1}, \frac{\sqrt{s^*} }{\alpha - 1} \right)$ with probability at least $1- \delta$.

\end{Proof}

\section {Proof of Theorem \ref{main-results} } \label{sec: appendix_main-results}

\begin{Proof} Our bounds respectively follow from Equations \eqref{fundamental-lasso} and \eqref{fundamental-group}.
    
\paragraph{Proof for Lasso: }
As a second consequence of Equation \eqref{fundamental-lasso}, it holds with probability at least $1 - \delta$:
\begin{equation} 
\begin{split}
\frac{1}{n} \| \B{X} \B{h} \|_2^2  
&\le \frac{\lambda}{\alpha} \left( \sqrt{k^*}  \| \B{h}_{S_0}  \|_2  +  \| \B{h}_{(S_0)^c}  \|_1 \right) + \lambda \| \B{h}_{S_0} \|_1 -\lambda \| \B{h}_{(S_0)^c}  \|_1\\
&\le \frac{\lambda}{\alpha} \sqrt{k^*}   \| \B{h}_{S_0} \|_{2} + \lambda \| \B{h}_{S_0} \|_{1} \\
&\le 2 \lambda \sqrt{k^*}   \| \B{h}_{S_0} \|_{2}
\le 2 \lambda \sqrt{k^*}   \| \B{h} \|_{2}, 
\end{split}
\end{equation}
where we have used Cauchy-Schwartz inequality on the $k^*$ sparse vector $\B{h}_{S_0} $.

\smallskip
\noindent
The cone condition proved in Theorem \ref{cone-condition} gives $\B{h}= \hat{\B{\beta}} _1- \B{\beta}^* \in \Lambda \left( S_0, \ \gamma_1^*=\frac{\alpha}{\alpha - 1},  \ \gamma_2^*=\frac{\sqrt{k^*}}{\alpha - 1} \right)$. We can then use the restricted eigenvalue condition defined in Assumption \ref{asu4}.1$(k^*, \gamma^*)$---where we define $\kappa^* = \kappa \left(k^*, \gamma_1^*, \gamma_2^*\right)$. It then holds with probability at least $1 - \delta$:
$$ \kappa^* \| \B{h} \|_2^2  \le  \frac{1}{n} \| \B{X} \B{h}\|_2^2 \le 2 \lambda \sqrt{k^*}   \| \B{h} \|_{2}.$$
By using that $\lambda=34 \alpha \sigma \sqrt{ n^{-1} \log(2pe/k^*) \log(1 / \delta)}$, we conclude that it holds with probability at least $1 - \delta$:
\begin{equation*}
\begin{split}
&\|\B{h}\|_2^2 \lesssim 
\left( \frac{ \alpha \sigma }{\kappa^*} \right)^2 \frac{ k^* \log\left( p/k^* \right) \log\left( 1/\delta \right) }{n}.
\end{split}
\end{equation*}

\paragraph{Proof for Group Lasso: }
Similarly, as a second consequence of Equation \eqref{fundamental-group}, it holds with probability at least $1 - \delta$:
\begin{equation} 
\begin{split}
\frac{1}{n} \| \B{X} \B{h} \|_2^2  
&\le \frac{\lambda_G}{\alpha} \sqrt{s^*}   \| \B{h}_{\mathcal{T}_0} \|_{2} + \lambda_G \sum_{g \in \mathcal{J}_0} \| \B{h}_g \|_{2} 
\le 2 \lambda_G \sqrt{s^*}   \| \B{h} \|_{2}, 
\end{split}
\end{equation}
where we have used Cauchy-Schwartz inequality to obtain: $\sum_{g \in \mathcal{J}_0} \| \B{h}_g \|_{2} \le  \sqrt{s^*} \| \B{h}_{\mathcal{T}_0} \|_{2}$

\smallskip
\noindent
The cone condition proved in Theorem \ref{cone-condition} gives $\B{h} = \hat{\B{\beta}}_{L1-L2}  - \B{\beta}^*\in \Omega \left( \mathcal{J}_0,  \epsilon_1^*=\frac{\alpha}{\alpha-1},  \epsilon_2^* = \frac{\sqrt{s^*} }{\alpha - 1}  \right)$. We can then use the restricted eigenvalue condition defined in Assumption \eqref{asu4}.2$(s^*, \epsilon^*)$---where we have defined $\kappa^* = \kappa \left(s^*, \epsilon_1^*, \epsilon_2^*\right)$. It then holds:
$$\kappa^* \| \B{h} \|_2^2  \le 2\lambda_G \sqrt{s^*}   \| \B{h} \|_{2}.$$
We conclude, by using the definition of $\lambda_G = 34 \alpha \sigma \sqrt{ n^{-1} \log(2Ge/s^*) \log(1/ \delta)} + 4 \alpha \sigma \sqrt{ n^{-1}\gamma (s^*) ^{-1} m^* }$, that it holds with probability at least $1 - \delta$:
\begin{equation*}
\begin{split}
&\|\B{h}\|_2^2 \lesssim 
\left( \frac{ \alpha \sigma }{\kappa^*} \right)^2 \frac{ s^* \log\left( G/s^* \right) \log\left( 1/\delta \right) + \gamma m^* }{n}.
\end{split}
\end{equation*}
\end{Proof}

\section {Proof of Corollary \ref{main-corollary} } \label{sec: appendix_main-corollary}

\begin{Proof}
    In order to derive the bound in expectation, we define the bounded random variable: 
    $$ Z =  \frac{{\kappa^*}^{2} }{\alpha^2 \sigma^2} \| \hat{\B{\beta}}  - \B{\beta}^*\|_2^2,$$
    where $\kappa^*$ depends upon  the regularization used. We fix $C_0 > 0$ such that $\forall \delta \in \left(0, \frac{1}{n} \right)$, it holds with probability at least $1-\delta$:
    \begin{align*} 
    \begin{split}
    Z &\le C_0  H_1  \log(1/\delta)  +  C_0 H_2,
    \end{split}
    \end{align*} 
    where $H_1 = n^{-1} k^* \log\left( p/k^* \right)$, $H_2=0$ for Lasso and $H_1 = n^{-1} s^* \log\left( G/s^* \right)$, $H_2 =  n^{-1} \gamma m^*$ for Group Lasso. It then holds $\forall t \ge t_0 = \log(2):$
    $$\mathbb{P}\left( Z/C_0 \ge H_1 t + H_2 \right) \le e^{-t}.$$
    Let $q_0 = H_1 t_0$, then $\forall q \ge q_0$:
    \begin{align*}
    \mathbb{P}\left( Z/C_0 \ge q + H_2\right) &\le \exp\left( - \frac{q}{H_1}  \right).
    \end{align*}
    As a consequence, by integration, we have:
    \begin{align} 
    \begin{split}
    \mathbb{E}(Z) &= \displaystyle \int_0^{+ \infty}  C_0\mathbb{P}\left( |Z| /C_0 \ge q \right)dq\\
    &\le \displaystyle \int_{H_2 + q_0}^{+ \infty}  C_0 \mathbb{P}\left( |Z| /C_0 \ge  q\right) dq + C_0 (H_2 + q_0) = \displaystyle \int_{q_0}^{+ \infty}  C_0 \mathbb{P}\left( |Z| /C_0 \ge  q + H_2\right) dq + C_0 (H_2 + q_0) \\
    &\le \displaystyle \int_{q_0}^{+ \infty}  C_0 \exp \left(-\frac{q}{H_1}   \right)dq +  C_0 H_2 + C_0 H_1 t_0 \\
    &\le C_0 H_1 \exp \left( -\frac{q_0}{H_1}  \right)  +  C_0 H_2 + C_0 H_1 \log(2)  \\
    &\le C_1 \left( H_1 + H_2 \right) \text{ where } C_1 = 2C_0 + \log(2).
    \end{split}
    \end{align}
    Consequently, we conclude that:
    $$\mathbb{E} \| \hat{\B{\beta}} - \B{\beta}^*  \|_2^2  \lesssim   \left( \frac{\alpha \sigma}{\kappa^*} \right)^2 \left( H_1 +  H_2\right),$$
    which, for the Lasso estimator, is equivalent to:
    \begin{equation*}
    \mathbb{E} \| \hat{\B{\beta}}_{1}  - \B{\beta}^*\|_2^2 \lesssim \left( \frac{ \alpha \sigma}{\kappa^* } \right)^2 \frac{k^* \log\left( p /k^* \right)}{n},
    \end{equation*}
    and, for the Group Lasso estimator, can be equivalently expressed as:
    \begin{equation*}
    \mathbb{E} \| \hat{\B{\beta}}_{L1-L2}  - \B{\beta}^*\|_2^2 \lesssim \left( \frac{\alpha \sigma}{\kappa^*} \right)^2 \frac{s^* \log\left( G / s^* \right) + \gamma m^*}{n}.
    \end{equation*}
\end{Proof}

\end{appendices}

\end{document}